\documentclass[conference]{IEEEtran}
\IEEEoverridecommandlockouts
\usepackage{cite}
\usepackage{hyperref}
\usepackage{amsmath,amssymb,amsfonts}
\usepackage{graphicx}
\usepackage{textcomp}
\usepackage{xcolor}
\usepackage{float}
\usepackage{array}
\usepackage[utf8]{inputenc}
\usepackage[T5]{fontenc}
\usepackage{blindtext}
\usepackage{mathtools}
\usepackage{multirow}
\usepackage{algorithm}
\usepackage{algpseudocode}
\usepackage{hyperref}
\usepackage{algorithmicx}
\usepackage{tabularx}
\usepackage{footmisc}
\usepackage{multicol}
\usepackage{authblk}
\usepackage{hyperref}
\usepackage{makecell}
\usepackage{longtable}
\usepackage{tabularx}
\usepackage{pgfplots}
\usepackage{threeparttable}
\usepackage[font={small},justification=centering]{caption}
\usepackage{array}

\makeatletter
\renewcommand\footnoterule{%
  \kern-3\p@
  \hrule\@width1\columnwidth
  \kern2.7\p@}
\makeatother

\newcolumntype{C}[1]{>{\centering\let\newline\\\arraybackslash\hspace{0pt}}m{#1}}

\begin{document}
\title{Comparison Between Traditional Machine Learning Models And Neural Network Models For Vietnamese Hate Speech Detection \\
}

\author[1,2,*]{Son T. Luu}
\author[1,2,†]{Hung P. Nguyen}
\author[1,2,*]{Kiet Van Nguyen}
\author[1,2,*]{Ngan Luu-Thuy Nguyen}
\affil[1]{University of Information Technology, Ho Chi Minh City, Vietnam}
\affil[2]{Vietnam National University, Ho Chi Minh City, Vietnam}

\affil[ ]{Email: *\textit {\{sonlt,kietnv,ngannlt\}@uit.edu.vn}, †\textit{\{17520068\}@gm.uit.edu.vn}}

\maketitle

\begin{abstract}
Hate-speech detection on social network language has become one of the main researching fields recently due to the spreading of social networks like Facebook and Twitter. In Vietnam, the threat of offensive and harassment cause bad impacts for online user.  The VLSP - Shared task about Hate Speech Detection on social networks showed many proposed approaches for detecting whatever comment is clean or not. However, this problem still needs further researching. Consequently, we compare traditional machine learning and deep learning on a large dataset about the user's comments on social network in Vietnamese and find out what is the advantage and disadvantage of each model by comparing their accuracy on F1-score, then we pick two models in which has highest accuracy in traditional machine learning models and deep neural models respectively. Next, we compare these two models capable of predicting the right label by referencing their confusion matrices and considering the advantages and disadvantages of each model. Finally, from the comparison result, we propose our ensemble method that concentrates the abilities of traditional methods and deep learning methods.
\end{abstract}

\begin{IEEEkeywords}
hate-speech detection, offensive, harassment, comments, data, machine learning, deep neural network, comparison, imbalance data, classification, social network
\end{IEEEkeywords}

\section{INTRODUCTION}
\label{section1}
Keep the online conversation-friendly is the important task of social network provider. With the vast development of social network users in Vietnam as well as the increment in E-commerce, automated classification of hate, offensive and fool comments is essential to keep the clean environment for online discussion. Facebook\footnote[1]{\url{https://www.facebook.com/communitystandards/objectionable_content}} and Google\footnote[2]{\url{https://www.youtube.com/about/policies/\#community-guidelines}} also have policies for banning and blocking offensive and racist comments.

Researching on classifying social network comments deals with common challenges in text processing and natural network processing such as long-range dependencies, misspelled words and unformal word. Social networks users often use acronyms, idiosyncratic and non-formal words thus this common language processing such as tokenization or sentiment analysis is not fixed with this data. Besides, users may use many special characters likes emoji icons (ideograms and smileys used in electronic contents like email or social posts) and hashtags. Proposed solutions for this problem are using the traditional machine learning combine with features extractions, using pre-trained words embedding with deep neural networks like CNN or LSTM and its variances\footnote[3]{\url{https://www.kaggle.com/c/jigsaw-toxic-comment-classification-challenge/notebooks}}. However, these methods are suffering from the lack of training data therefore they often perform inaccuracy when dealing with real-world data. Hence it is important to survey different models on the dataset to find out which is the best choice. 

The Hate-speech detection data is provided by the 2019 VLSP Shared Task - Hate Speech Detection on Social network\footnote[4]{\url{https://www.aivivn.com/contests/8}}. The task is organized as multi-class classification task based on the dataset contains 20,345 comments, each comment is labeled by one of three classes: CLEAN, OFFENSIVE and CLEAN \cite{sonvx2019}. The purpose of this task is building a classification model which can classify an item to one of three class: CLEAN, OFFENSIVE or HATE. Based on the dataset about Hate-speech detection for good in \cite{sonvx2019}, we implemented different classifiers include traditional classifiers such as Support Vector Machine (SVM) and Logistic Regression and deep neural network classifiers such as GRU and Text-CNN. By comparing the accuracy of each model, we can find advantages and disadvantages of each model on this problem, then we will suggest a method that combining the advantage of models to build an ensemble model that suitable for predicting hate-speech data in the future. 

In this paper, we have three main contributions as described follows.
\begin{itemize}
\item First, we conducted experiments on four different classifiers for the dataset. Each classifier is performed on the same training dataset and get the results by F1- score index.
\item Second, according to the experimental result given by each model we analyze which model is suitable for each class, then we find advantages and disadvantages of the model.
\item Third, we suggest next researching in the future for Hate-speech detection problem in Vietnamese language.
\end{itemize}

The content of the paper is structured as follow. Section \ref{section2} lists the related works on Hatespeech detection. Section \ref{section3} takes an overview on VLSP2019 's Hatespeech dataset. Section \ref{section4} presents our implementation for solving the Hatespeech detection problem on the dataset. Section \ref{section5} illustrates experiments and results on the VLSP 's dataset. Finally, Section \ref{section6} concludes the paper and discusses future works.
\section{RELATED WORKS}
\label{section2}
Waseem and Hovy (2016) presented a list of criteria based on critical race theory to identify racist and slurs. They also use the Logistic regression model with different features include char n-grams, word n-grams, gender, gender and location and finally evaluate the accuracy when using each feature \cite{waseem2016}.

Davidson et al. (2017) using the lexical method and multi-class classifier to classify hate speech with offensive speech. They also built the Hatebase dataset which contains data from nearly 24K Twitter's user labeled tweets. However, they find that it is difficult to identify between Hate and Offensive by a lexical method \cite{davidson2017}.

Gaydhani et al. (2018) solve the hate speech detection problem on Twitter platform using dataset that combine three other different datasets. Then they use Logistic regression with n-gram range 1 to 3 and L2 normalization of TF-IDF value features and get the accuracy 95.6\% \cite{gaydhani2018}.

Fortuna and Nunes (2018) has studied about the motivation of reseaching on Hatespeech dection problem and gave the definition of hate speech based on several sources such as European Union Commission, International minorities associations (ILGA), scientific papers and social networks. Besides, the authors also took surveys on how to approach the Hatespeech detection problem as well as the current research challenges and discussed about future works \cite{fortuna2018}.

Martin et al. (2018) using the SVM (Support Vector Machine) with features using Weka Ranker Search with InfoGainAttributeEval as an attribute evaluator on the Hatebase dataset \cite{davidson2017}. The result with SVM is 80.56\%, which is the best result \cite{martin2018}.

Georgakopoulos et al. (2018) apply the CNN model with word2vec embedding (CNNfit) and CNN model with randomly initialized words and tuned during training (CNNrand) on Kaggle Toxic Comment Challenge dataset\footnote[1]{\url{https://www.kaggle.com/c/jigsaw-toxic-comment-classification-challenge}}. The accuracy for CNNfit and CNNrand is 91.2\% and 89.5\% respectively \cite{georgakopoulos2018}. 

Sharma and Patel (2018) have used CNN model and LSTM model for classifying toxic comments to 6 different labels. The result shows that LSTM model performing better than CNN (98.77\% for LSTM over 98.04\% for CNN) \cite{sharma2018}.

Ibrahim et al. (2018) proposed three data augmentation techniques to solve the imbalance data in Wikipedia 's talk page edit dataset provided in Kaggle Toxic Comment Classification competition. Besides authors also proposed an ensemble method that combines three models: CNN, LSTM, and Bi-GRU \cite{ibrahim2018}. 

The SemEval 2019 has organized the shared task on Multilingual Detection of Hate (English and Spanish). The challenge contains two tasks: The first task (Task A) focus on detecting whether a given tweet contains hatred about immigrants or women, the second task (Task B) focus on classifying hateful tweets as aggressive or not aggressive. The result showed that English gave 0.378 macro F1 on task A and 0.553 macro F1 on task B, Spanish gave 0.701 macro F1 on task A and 0.734 macro F1 on task B \cite{zhang2019}.

According to the 2019 VLSP workshop about Hate Speech Detection on Social network Shared task \cite{sonvx2019}, the organizer has chosen the top 5 teams with the highest score based on their proposed model 's accuracy in macro F1-score: 
\begin{itemize}
\item Pham et al. has proposed an ensemble method and archived the first position on the contest table with 67.76\% for public test and 61,97\% for private test \cite{quang2019}.
\item Nguyen et al. proposed their ensemble method and gain the result with 73\% for public test and 58.4\% for private test \cite{binh2019}.
\item Dang et al. used ensemble method with replied features extraction strategy and received 70.852\% on public test and 58.883\% for private test \cite{thin2019}.
\item Do et al. used Bi-LSTM for building the model and get accuracy with 71.43\% on F1-score \cite{hang2019}.
\item Huynh et al. implemented learning method based on Bi-GRU-LSTM-CNN classifier. The result is 70.576\% of F1-score \cite{tin2019}. This model was also successfully implemented on the problem of the job classification \cite{van2019job}.
\end{itemize}
\section{THE DATASET}
\label{section3}
The Hate-speech dataset was provided by the VLSP shared task \cite{sonvx2019}. The dataset contains 20,345 comments or posts from Facebook. Each comment or post is annotated with one of three labels: CLEAN, OFFENSIVE and HATE. Table \ref{table1} describes the sample data from the dataset:

\begin{table}[H]
\caption{SAMPLE DATA FROM DATASET ON THREE CLASSES}
\begin{center}
\begin{tabular}{|c|c|c|}
\hline
\textbf{No. } & \textbf{Comments / Posts content} & \textbf{Label} \\
\hline
1 & \makecell{Cho xíu nhạc đi a \\ (\textbf{English:} Give me some music)} & CLEAN \\
\hline
2 & \makecell{Tao ghét vl. \\ (\textbf{English:} I fucking sulky this )} & OFFENSIVE \\
\hline
3 & \makecell{Lấy cm của bạn này ko bọn đầu bò lại khó hiểu\\ rồi đi chửi người khác ngu \\ (\textbf{English:} See this guy 's \\comment your damn foolist !)} & HATE \\
\hline
\end{tabular}
\end{center}
\label{table1}
\end{table}

According to \cite{sonvx2019}, the meaning of 3 labels is described as below: 
\begin{itemize}
\item \textbf{HATE}: a post or comment is identified as HATE if it contains profane and offensive words or attack an individual or groups based on their characteristics or encourage hatred.
\item \textbf{OFFENSIVE}: a post or comment contains offensive words but does not attack anyone or group person based on their characteristics.
\item \textbf{CLEAN}: Does not contains offensive words or hate language.
\end{itemize}

The number of posts/comments, average word length and vocabulary size on each label is described in the table below:

\begin{table}[H]
\caption{STATISTICS OF THE VLSP HATE SPEECH DETECTION DATASET}
\begin{center}
\begin{tabular}{|c|c|c|c|c|}
\hline
Label & \textbf{CLEAN} & \textbf{HATE} & \textbf{OFFENSIVE} & \textbf{All} \\
\hline
Number of comments & 18,614 & 709 & 1,022 & 20,345\\
\hline
Average word length & 18.69 & 20.46 & 9.35 & 18.28 \\
\hline
Vocabulary size & 347,949 & 14,513 & 9,556 & 372,018 \\
\hline
\end{tabular}
\end{center}
\label{table2}
\end{table}

In Table \ref{table2}, the vocabulary size is calculated by the sum of amount of words by each label. Average word length is calculated by dividing vocabulary size by number of comments on each label. 

According to \ref{table2}, the dataset is unbalanced because the number of comments of CLEAN label is significantly much more than the two remain labels. HATE label is the least, but have the highest average word length. The OFFENSIVE label 's average word length is the lowest. The amount of comments and average word length are illustrated as below.

\begin{figure}[H]
\begin{minipage}[t]{.24\textwidth}
    \definecolor{grey}{HTML}{b4b4a9}
    \begin{tikzpicture}[scale=0.58]
        \begin{axis}[
            ybar,
            enlarge y limits={0.15,upper},
            enlarge x limits=0.45,
            symbolic x coords={CLEAN,HATE,OFFENSIVE},
            xtick=data,
            ymin = 0, ymax = 19000,
            nodes near coords,
            nodes near coords align={vertical},
    	    ylabel near ticks,
        ]
        \addplot[black, fill=grey] coordinates {(CLEAN,18614) (HATE,709) (OFFENSIVE,1022)};
        \end{axis}
    \end{tikzpicture}
    \caption{\textit{Number of comments on three labels}} 
    \label{figures1}
\end{minipage}
\begin{minipage}[t]{.2\textwidth}
    \definecolor{grey}{HTML}{b4b4a9}
    \begin{tikzpicture}[scale=0.58]
        \begin{axis}[
            ybar,
            enlarge y limits={0.15,upper},
            enlarge x limits=0.45,
            symbolic x coords={CLEAN,HATE,OFFENSIVE},
            xtick=data,
            ymin = 0, ymax = 20,
            nodes near coords,
            nodes near coords align={vertical},
    	    ylabel near ticks,
        ]
        \addplot[black, fill=grey] coordinates {(CLEAN,18.69) (HATE,20.46) (OFFENSIVE,9.35)};
        \end{axis}
    \end{tikzpicture}
    \small
    \caption{\textit{Average word length of three labels}} 
    \label{figures2}
\end{minipage}
\end{figure}

Also, the comments and posts do not use formal language: users tend to use various of emotion icons (emoji), hashtags, attached links or even use acronyms and slangs, which is hard to use the common NLP 's processing techniques like word-tokenization and word segmentation. 

\section{THE METHODOLOGIES}
\label{section4}
In this section, we implement four methods that often use in text classification. The goal of those methods is trying to minimize the error between predicted labels for new comments or posts with the real label that the comments or posts belong to. Besides, we will use both traditional machine learning models and deep neural models for this problem to compare their efficiency. 

\subsection{Word embedding}
Word embedding is a feature learning technique in Natural Language Processing (NLP) by mapping words or phrases from the set of vocabularies to a vector of real numbers. The distributed representations of words, which called word embedding help to improve the accuracy of various natural language models \cite{hoang2017}. Thai-Hoang Pham and Phuong Le-Hong used recurrent neural network models with pre-trained word embeddings as input gained the first-rank in NER shared task organized by VLSP \cite{hoang2017}. 

In this paper, we use the Vietnamese word embedding\footnote[1]{\url{https://fasttext.cc/docs/en/crawl-vectors.html}} provided by Edouard Grave et al. This word embedding is trained on Common Craw and Wikipedia using CBOW with position-weights and in dimension 300 with character n-grams \cite{grave2018}. This pre-trained embedding model will be used as an embedding layer in two deep neural models training.

\subsection{Traditional models}
\textbf{Logistic regression (LR):} This one of the basic and famous algorithms for classification, especially binary classification. In text classification, it requires the manual feature extractions. In this paper, we use the Logistic regression with 2 features: 
\begin{enumerate}
\item TF-IDF with word analyzer, used stop-words\footnote[2]{\url{https://github.com/NguyenVanHieuBlog/vietnamese-stopwords/blob/master/stopwords.txt} \label{footnote2}} and n-gram range is (1,3) and maximum features is 20,000.
\item TF-IDF with char analyzer, n-gram range is (3,6), used stop-words\footref{footnote2} and maximum-features is 40,000.
\end{enumerate}

\textbf{Support Vector Machine (SVM:} This is a popular machine learning method for classification, regression, and other learning tasks. In LIBSVM \cite{chang2011}, the SVC (Support vector classification) kernel, which can use for two classes and multi-classes classification. Our experiment uses SVM with linear classifier and uses the same features extractor as Logistic regression: TF-IDF with word analyzer and TF-IDF with char n- gram analyzer, n-gram range is (3,6).

\subsection{Deep neural network models}
\textbf{Text-CNN:} Convolutional neural network (CNN) is a multistage Neural network architecture developed for classification. By using convolutional layers, it can detect combination features \cite{georgakopoulos2018,ho2019emotion}. In our experiments, we use 4 convolutional layers with 32 filters for each layer.

\textbf{GRU:} Gate Recurrent Unit (GRU) is a kind of Recurrent Neural Network (RNN) model and is a variance of LSTM model. Unlike LSTM with three gates: input gate, output gate and forget gate, Gate recurrent unit has only two gates: reset gate and update gate thus it is less complicated than LSTM and training faster than LSTM model \cite{zhang-lei2018}. 
\section{EXPERIMENTS}
\label{section5}
\subsection{Data preparation}
Based on the dataset described in Section \ref{section3}, we first apply pre-processing steps as described below to make the dataset clean: 
\begin{itemize}
\item Removing special characters such as emoji icons and hashtags.
\item Removing the URL links and digits.
\item Changing words to lower-case.
\item Tokenizing the document to words by using Pyvi tools\footnote[3]{\url{https://pypi.org/project/pyvi}}
\end{itemize}

After cleaned the dataset, we applied two traditional models and two deep neural models respectively. On each model, we divided the dataset into the training set and testing set with ratio 8:2, which means 80 percent for training data and 20 percent for testing data. 

Then we split our data into 5 folds, each fold has the same proportions of three classes. On each fold, we train the model on training data and use the testing data for validating the accuracy of the predicted label. The accuracy is computed by using the F1-score macro metric \cite{asch2013}.

Finally, we calculated the accuracy of the whole training phase by calculated the means of accuracy on 5 folds, and this is also the result of the model. 

In addition, we also computed the confusion matrix on each model. The value of the confusion matrix on each model is the average value of 5 confusion matrices according to 5 folds. On each gold label, we computed the percentage of each correspond predicted label.

\subsection{Experimenal configuration}
\textbf{Logistic regression (LR):} We use Logistic regression with C = 1.0 and balanced class weight.

\textbf{Support Vector Machine (SVM):} We use the SVC (Support Vector Classification) with linear kernel and C = 1.0 (C is Regularization parameter).

\textbf{Text-CNN:} We build the Text-CNN model with an embedding layer and 4 Conv layers. Embedding layer has embedding size 300 and maximum features 11,221. Each conv layer has 32 filters and the sizes of the filter is 1, 2, 3 and 5 respectively. The conv layer activation is ELU (Exponential Linear Unit). The output layer is Dense 3 with sigmoid activation function, corresponding to three classes.

\textbf{Gate Recurrent Unit (GRU):} This model has an embedding layer with embedding size 300 and the Bidirectional RNN layer is GRU with 80 units. The output layer is Dense 3 with sigmoid activation function, corresponding to three classes.

\subsection{Results and discussion}
The table below described the accuracy of four models on Hate-speech dataset:
\begin{table}[H]
\caption{RESULTS ON HATE SPEECH DATA}
\begin{center}
\begin{tabular}{|c|c|c|}
\hline
\multicolumn{2}{|c|}{\textbf{Model}} & \textbf{F1-macro score} \\
\hline
\multirow{ 2}{*}{\makecell{Traditional \\ machine learning}} & \textbf{SVM} & \textbf{65.10} \\
\cline{2-3}
& Logistic regression & 64.58 \\
\hline
\multirow{ 2}{*}{\makecell{Deep neural network \\ models}} & \textbf{Text-CNN} & \textbf{83.04} \\
\cline{2-3}
& GRU & 80.12 \\
\hline
\end{tabular}
\end{center}
\label{table3}
\end{table}

Based on Table \ref{table3}, we can infer that Support Vector Machine giving the best result among traditional models (65.10\%) and Text-CNN giving the best result on deep neural network models (83.04\%). According to those two models, Table \ref{table4} and Table \ref{table5} describe the confusion matrix of SVM and Text-CNN respectively. 

\begin{table}[H]
\caption{CONFUSION MATRIX OF EXPERIMENTAL RESULTS ON  SVM}
\begin{center}
\begin{threeparttable}
\begin{tabular}{|c|c|c|c|}
\hline
& \textbf{Predicted - 0} & \textbf{Predicted - 1} & \textbf{Predicted - 2} \\
\hline
\textbf{Gold - 0} & \makecell{\textit{3694.2} \\ \textit{(99.23\%)}} & \makecell{17.0 \\ (0.45\%)} & \makecell{11.8 \\ (0.31\%)} \\
\hline
\textbf{Gold - 1} & \makecell{130.0 \\ (63.72\%)} & \makecell{\textit{57.8} \\ \textit{(28.33\%)}} & \makecell{16.2 \\ (7.94\%)} \\
\hline
\textbf{Gold - 2} & \makecell{55.8 \\ (39.29\%)} & \makecell{15.2 \\ (10.70\%)} & \makecell{\textit{71.0} \\ \textit{(50.00\%)}} \\
\hline
\end{tabular}
\begin{tablenotes}
    \small
    \item \textit{Note: 0 - CLEAN, 1 - OFFENSIVE, 2 - HATE}
\end{tablenotes}
\end{threeparttable}
\end{center}

\label{table4}
\end{table}

\begin{table}[H]
\caption{CONFUSION MATRIX OF EXPERIMENTAL RESULTS ON  TEXT-CNN}
\begin{center}
\begin{threeparttable}
\begin{tabular}{|c|c|c|c|}
\hline
& \textbf{Predicted - 0} & \textbf{Predicted - 1} & \textbf{Predicted - 2} \\
\hline
\textbf{Gold - 0} & \makecell{\textit{3693.8} \\ \textit{(99.21\%)}} & \makecell{22.2 \\ (0.59\%)} & \makecell{7.0 \\ (0.19\%)} \\
\hline
\textbf{Gold - 1} & \makecell{61.0 \\ (29.90\%)} & \makecell{\textit{133.8} \\ \textit{(65.58\%)}} & \makecell{9.2 \\ (4.51\%)} \\
\hline
\textbf{Gold - 2} & \makecell{29.2 \\ (20.56\%)} & \makecell{7.8 \\ (5.49\%)} & \makecell{\textit{105.0} \\ \textit{(73.94\%)}} \\
\hline
\end{tabular}
\begin{tablenotes}
    \small
    \item \textit{Note: 0 - CLEAN, 1 - OFFENSIVE, 2 - HATE}
\end{tablenotes}
\end{threeparttable}
\end{center}

\label{table5}
\end{table}

According to table \ref{table4} and table \ref{table5}, assumes that:
\begin{itemize}
\item \textbf{Label-0:} CLEAN label.
\item \textbf{Label-1:} OFFENSIVE label.
\item \textbf{Label-2:} HATE label.
\item \textbf{True-0:} Observation is 0, predicted as 0.
\item \textbf{True-1:} Observation is 1, predicted as 1.
\item \textbf{True-2:} Observation is 2, predicted as 2.
\item \textbf{False-0-1:} Observation is 0, predicted as 1.
\item \textbf{False-0-2:} Observation is 0, predicted as 2.
\item \textbf{False-1-0:} Observation is 1, predicted as 0.
\item \textbf{False-1-2:} Observation is 1, predicted as 2.
\item \textbf{False-2-0:} Observation is 2, predicted as 0.
\item \textbf{False-2-1:} Observation is 2, predicted as 1.
\end{itemize}

Based on results from Table \ref{table4} and Table \ref{table5} with assumptions as below, we compared the ability of predicted right label of two models: SVM and Text-CNN.

According to Table \ref{table4}, SVM gave a very high result on \textbf{Label-0}, which takes 99.23\% on \textbf{True-0}. The mistaken prediction of SVM is very low (only 0.45\% for \textbf{False-0-1} and 0.31\% for \textbf{False-0-2}). For Text-CNN, according to Table \ref{table5}, the \textbf{True-0} is also very high with 99.21\%, but lower than SVM a little bit. In general, both SVM and Text-CNN give the best result on predicting \textbf{Label-0}, and SVM seems better than Text-CNN on predicting \textbf{Label-0}.

Nevertheless, on \textbf{Label-1}, SVM result is worse than Text-CNN. According to Table \ref{table4}, many of comments of \textbf{Label-1} items was predicted as \textbf{Label-0} by SVM (\textbf{False-1-0 of SVM is 62.35\%}), while the right prediction \textbf{True-1} is lower (\textbf{True-1} of SVM is 28.33\%). This means SVM tends to predict items of \textbf{Label-1} as \textbf{Label-0}. In contrast, Text-CNN result on \textbf{Label-1} is optimistic. The \textbf{True-1} of Text-CNN is higher than SVM (65.58\% for Text-CNN over 28.33\% for SVM). Besides, the mistaken predictions of Text-CNN on \textbf{Label-1} is also very low (29.90\% for \textbf{False-1-0} and 4.51\% for \textbf{False-1-2}). It can be inferred that, SVM seems to make wrong prediction with items on \textbf{Label-1}, specifically predicts \textbf{Label-1} items to \textbf{Label-0}. Text-CNN, instead, give the good result on predicting items belong to \textbf{Label-1}.

Moreover, on \textbf{Label-2}, according to Table \ref{table4} and Table \ref{table5}, the \textbf{True-2} percentage of Text-CNN is higher than SVM (73.94\% for Text-CNN and 50.00\% for SVM). The mistaken prediction \textbf{False-2-0} and \textbf{False-2-1} of Text-CNN are also lower than SVM. Therefore, on \textbf{Label-2}, Text-CNN again predicts better than SVM. In addition, SVM results on \textbf{Label-2} is better than \textbf{Label-1}. The \textbf{True-2} of SVM is higher than \textbf{True-1} (50.00\% for \textbf{True-2} over 28.33\% for \textbf{True-1}). Besides, the wrong predictions on \textbf{Label-2} of SVM is also lower than wrong predictions on \textbf{Label-1}. Thus, for \textbf{Label-2} items, SVM predicts better than \textbf{Label-1} items.

Furthermore, according to Table \ref{table4} and Table \ref{table5}, the \textbf{False-1-2} percentage and \textbf{False-2-1} percentage of Text-CNN are lower than SVM. (4.51\% for Text-CNN over 7.94\% for SVM on \textbf{False-1-2} and 5.49\% for Text-CNN over 10.07\% on \textbf{False-2-1}). As result, Text-CNN predicts \textbf{Label-1} and \textbf{Label-2} items better than SVM. 

In addition, according to Table \ref{table4} and \ref{table5}, on \textbf{Label-1}, the \textbf{False-1-0} is higher than \textbf{False-1-2} on both SVM and Text-CNN. It can be inferred that, items in \textbf{Label-1} are mostly predicted as \textbf{Label-0}. Items in \textbf{Label-2} are similar. The \textbf{False-2-0} is higher than \textbf{False-2-1} on both SVM and Text-CNN, thus most of items in \textbf{Label2} are predicted as \textbf{Label-0}. In contrast, the mistaken between \textbf{Label-1} and \textbf{Label-2} is trivial (The \textbf{False-1-2} and \textbf{False-2-1} are low). This was caused by the imbalance in the dataset as described in Section \ref{section3}. Figure \ref{figure4} and Figure \ref{figure5} illustrate the false predictions percentage on \textbf{Label-1} and \textbf{Label-2} respectively on both SVM and Text-CNN.

\begin{figure}[H]
\begin{minipage}[t]{.24\textwidth}
    \definecolor{grey}{HTML}{b4b4a9}
    \begin{tikzpicture}[scale=0.58]
        \begin{axis}[
            ybar,
            enlarge x limits=0.45,
            symbolic x coords={False-1-0, False-1-2},
            xtick=data,
            ymin = 0, ymax = 100,
    	    ylabel near ticks,
        ]
            \addplot[black, fill=grey] coordinates {(False-1-0,63.72) (False-1-2,7.94)};
            \addplot[black, fill=white] coordinates {(False-1-0,29.90) (False-1-2,4.51)};
            
            \legend{SVM, Text-CNN}
        \end{axis}
    \end{tikzpicture}
    \caption{\textit{False predictions on Label-1}} 
    \label{figure4}
\end{minipage}
\begin{minipage}[t]{.2\textwidth}
    \definecolor{grey}{HTML}{b4b4a9}
    \begin{tikzpicture}[scale=0.58]
        \begin{axis}[
            ybar,
            enlarge x limits=0.45,
            symbolic x coords={False-2-0, False-2-1},
            xtick=data,
            ymin = 0, ymax = 100,
    	    ylabel near ticks,
        ]
            \addplot[black, fill=grey] coordinates {(False-2-0,39.29) (False-2-1,10.70)};
            \addplot[black, fill=white] coordinates {(False-2-0,20.56) (False-2-1,5.49)};
            
            \legend{SVM, Text-CNN}
        \end{axis}
    \end{tikzpicture}
    \small
    \caption{\textit{False predictions on Label-2}} 
    \label{figure5}
\end{minipage}
\end{figure}

In addition, Table \ref{table6} and Table \ref{table7} describe the confusion matrix for Gate Recurrent Unit (GRU) and Logistic regression (LR) respectively.

\begin{table}[H]
\caption{CONFUSION MATRIX OF EXPERIMENTAL RESULTS ON GRU}
\begin{center}
\begin{threeparttable}
\begin{tabular}{|c|c|c|c|}
\hline
& \textbf{Predicted - 0} & \textbf{Predicted - 1} & \textbf{Predicted - 2} \\
\hline
\textbf{Gold - 0} & \makecell{\textit{3687.0} \\ \textit{(99.03\%)}} & \makecell{24.6 \\ (0.66\%)} & \makecell{11.4 \\ (0.30\%)} \\
\hline
\textbf{Gold - 1} & \makecell{61.0 \\ (29.90\%)} & \makecell{\textit{127.2} \\ \textit{(62.35\%)}} & \makecell{15.8 \\ (7.74\%)} \\
\hline
\textbf{Gold - 2} & \makecell{26.4 \\ (18.59\%)} & \makecell{19.8 \\ (13.94\%)} & \makecell{\textit{95.8} \\ \textit{(67.46\%)}} \\
\hline
\end{tabular}
\begin{tablenotes}
    \small
    \item \textit{Note: 0 - CLEAN, 1 - OFFENSIVE, 2 - HATE}
\end{tablenotes}
\end{threeparttable}
\end{center}
\label{table6}
\end{table}

\begin{table}[H]
\caption{CONFUSION MATRIX OF EXPERIMENTAL RESULTS ON LOGISTIC REGRESSION}
\begin{center}
\begin{threeparttable}
\begin{tabular}{|c|c|c|c|}
\hline
& \textbf{Predicted - 0} & \textbf{Predicted - 1} & \textbf{Predicted - 2} \\
\hline
\textbf{Gold - 0} & \makecell{\textit{3555.4} \\ \textit{(95.49\%)}} & \makecell{126.0 \\ (3.38\%)} & \makecell{41.6 \\ (1.11\%)} \\
\hline
\textbf{Gold - 1} & \makecell{84.0 \\ (41.17\%)} & \makecell{\textit{91.2} \\ \textit{(44.70\%)}} & \makecell{28.8 \\ (14.11\%)} \\
\hline
\textbf{Gold - 2} & \makecell{28.2 \\ (19.85\%)} & \makecell{30.0 \\ (21.12\%)} & \makecell{\textit{83.8} \\ \textit{(59.01\%)}} \\
\hline
\end{tabular}
\begin{tablenotes}
    \small
    \item \textit{Note: 0 - CLEAN, 1 - OFFENSIVE, 2 - HATE}
\end{tablenotes}
\end{threeparttable}
\end{center}

\label{table7}
\end{table}

According to \ref{table7} and table \ref{table4}, the \textbf{True-1} of SVM is worse than Logistic regression (SVM give only 28.33\% while logistic regression give 44.70\%). Besides, the \textbf{False-1-0} of Logistic regression is also lower than SVM (Logistic regression is 41.17\% and SVM is 63.72\%). This means, despite of giving lower result than SVM, Logistic regression having lower inaccurate predictions between \textbf{Label-1} and \textbf{Label-0} than SVM. 

Besides, as shown in \ref{table6} and \ref{table5}, GRU 's figures is lower than Text-CNN, but the proportion on three labels is seem similar to Text-CNN. 

Overall, deep neural models giving better result than traditional models. On traditional models, SVM giving the best result, but worse than Logistic regression on classifying between \textbf{Label-1} and \textbf{Label-0}. In deep neural models, Text-CNN give the best result on predicting of three labels and thus better than GRU. In brief, Text-CNN is a good model for classifying Hate speech texts.

\subsection{Error Analysis}

\begin{table}[H]
\caption{EXAMPLES OF CLASSIFICATION ERROR}
\begin{center}
\begin{threeparttable}
\begin{tabular}{|C{2em}|c|C{2.5em}|C{2.5em}|}
\hline
\textbf{No.} & \textbf{Example comment} & \textbf{True label} & \textbf{False label} \\
\hline
1 & \makecell{Dỗi vl \\ (\textbf{English:} How fucking sulky I feel now)} & 1 & 0 \\
\hline
2 & \makecell{Tội anh vl \\ (\textbf{English:} How fucking sulky I feel for you)} & 1 & 0 \\
\hline
3 & \makecell{Thế không tha thì lại bảo xấu như chó \\ (\textbf{English:} If I don't forgive you, \\then you will say I am an asshole)} & 1 & 2 \\
\hline
4 & \makecell{ad trẻ trâu vl \\ (\textbf{English:} Admin is being a scumbag)} & 2 & 0 \\
\hline
5 & \makecell{phò vl \\ (\textbf{English:} Such a slut)} & 2 & 0 \\
\hline
6 & \makecell{mày khôn lắm thằng ngu ạ \\ (\textbf{English:} You so smart, dumbass)} & 2 & 1 \\
\hline
\end{tabular}
\end{threeparttable}
\begin{tablenotes}
    \small
    \item \textit{Note: 0 - CLEAN, 1 - OFFENSIVE, 2 - HATE}
\end{tablenotes}
\end{center}
\label{table8}
\end{table}

Table \ref{table8} shows examples of classifications error. From the table, we can see that, many of comments that contains the words such as: "vl" (English: fucky sulky) was predicted as \textbf {Label-0} instead of \textbf{Label-1} or \textbf{Label-2}. Those words in Vietnamese are categorized as profanity words. Their appearance in user's comments can make these comments become offensive or hateful.  

After further observation to the dataset, we found that, many of comments in \textbf{Label-1} and \textbf{Label-2} contains the Vietnamese profanity words such as: "vl", "vkl", "dcm". Furthermore, the comments in \textbf{Label-2} not only have profanity words but also contains the Vietnamese personal pronouns such as: "mày", "thằng", "con". These are potential features for predicting offensive and hateful comments. The next study should pay more attention to these profanity words and personal pronouns.
\section{CONCLUSION AND FUTURE WORKS}
\label{section6}
In this paper, based on the Vietnamese Hate speech detection dataset provided by the 2019 VLSP shared task, we conducted different experiments on four models, two models belong to traditional machine learning models and two models belong to deep neural network models. On deep neural network model, we use the Vietnamese word embedding provided by Edouard Grave et al. From the results gained by F1-scores, we compare their ability of classifying on three classes by looking at their confusion matrix. The result is deep neural models are better than traditional models and Text-CNN is the good model for classifying hate speech texts with accuracy 83.04\% in F1-macro score. SVM model has accuracy 65.10\%, higher than Logistic regression, but is weak on classifying between OFFENSIVE label and CLEAN label. Besides, the lack of data in OFFENSIVE label and HATE label impact to the prediction models.

In addition, based on the error analysis, we found that profanity words and personal pronouns influenced to determine offensive and hateful comments. Thus next study should focus on those features to increase the right prediction.  

After all, the Hate speech detection problem on Vietnamese is a challenging problem for keeping a clean environment for discussion on the Social network. Classification content on social networks is not easy because of free-style writing, using acronyms, informal words, and slangs. For the future researching about hate speech detection in Vietnamese language, we propose three improvements:
\begin{enumerate}
\item Constructing a dictionary about profanity and foul words in Vietnamese. Then we build a word embedding on Vietnamese hate speech social media texts. 
\item Applying data augmentation techniques for solving the imbalance in the VSLP's dataset, especially in OFFENSIVE and HATE labels.
\item Contributing to the VLSP' dataset by collecting and annotating new data.
\end{enumerate}

\section*{Acknowledgment}
We would like to give our great thanks to the 2019 VLSP Shared task organizers for providing a very valuable dataset for our experiments. This research is funded by University of Information Technology, Vietnam National University HoChiMinh City under grant number D1-2020-02.

\bibliographystyle{IEEEtran}
\bibliography{reference}

\end{document}